\documentclass{article}
\usepackage{spconf}
\usepackage{amsmath}
\usepackage{graphicx}
\usepackage[table]{xcolor} 
\usepackage{enumitem}
\setlist{nosep, leftmargin=14pt}
\usepackage{amsfonts}  
\usepackage{bm}
\usepackage{hyperref}
\usepackage{booktabs}  
\usepackage{multirow}  
\usepackage{bbm}  
\usepackage{array}  
\usepackage{bbding}  
\usepackage{makecell}  
\usepackage{subcaption}

\newcommand{\realcolon}[1]{REAL-Colon}

\newcolumntype{L}[1]{>{\raggedright\let\newline\\\arraybackslash\hspace{0pt}}m{#1}}
\newcolumntype{C}[1]{>{\centering\let\newline\\\arraybackslash\hspace{0pt}}m{#1}}
\newcolumntype{R}[1]{>{\raggedleft\let\newline\\\arraybackslash\hspace{0pt}}m{#1}}

\title{Towards Polyp Counting In Full-Procedure Colonoscopy Videos}
\name{Luca Parolari$^{\dagger}$\Envelope \qquad Andrea Cherubini$^{\ddagger\mathsection
}$ \qquad Lamberto Ballan$^{\dagger}$ \qquad Carlo Biffi$^{\ddagger}$}

\address{$^{\dagger}$ Department of Mathematics, University of Padova, Padova, Italy \\
$^{\ddagger}$ Cosmo Intelligent Medical Devices, Dublin, Ireland \\
$^{\mathsection}$ Milan Center for Neuroscience, University of Milano–Bicocca, Milan, Italy\\
\Envelope luca.parolari@phd.unipd.it
}
\begin{document}
%
\maketitle
\begin{abstract}
Automated colonoscopy reporting holds great potential for enhancing quality control and improving cost-effectiveness of colonoscopy procedures.
A major challenge lies in the automated identification, tracking, and re-association (ReID) of polyps tracklets across full-procedure colonoscopy videos.
This is essential for precise polyp counting and enables automated computation of key quality metrics, such as Adenoma Detection Rate (ADR) and Polyps Per Colonoscopy (PPC).
However, polyp ReID is challenging due to variations in polyp appearance, frequent disappearance from the field of view, and occlusions.
In this work, we leverage the REAL-Colon dataset, the first open-access dataset providing full-procedure videos, to define tasks, data splits and metrics for the problem of automatically count polyps in full-procedure videos, establishing an open-access framework.
We re-implement previously proposed SimCLR-based methods for learning representations of polyp tracklets, both single-frame and multi-view, and adapt them to the polyp counting task.
We then propose an Affinity Propagation-based clustering method to further improve ReID based on these learned representations, ultimately enhancing polyp counting.
Our approach achieves state-of-the-art performance, with a polyp fragmentation rate of 6.30 and a false positive rate (FPR) below 5\% on the REAL-Colon dataset. We release code at \href{https://github.com/lparolari/towards-polyp-counting}{https://github.com/lparolari/towards-polyp-counting}.
\end{abstract}
\section{Introduction}
\label{sec:intro}
Colonoscopy helps prevent colorectal cancer by detecting and removing adenomatous polyps. Recent advancements in AI-assisted colonoscopy systems have shown potential to improve adenoma detection rates (ADR) and contribute to the standardization of colonoscopy procedures~\cite{berzin_position_2020, spadaccini_computer-aided_2021, aizenman2024assessing, biffi_novel_2022}. Next-generation AI-assisted systems are now expected to generate automated reports, listing all detected polyps
and computing key quality metrics, including the ADR and Polyps Per Colonoscopy (PPC)~\cite{tavanapong2022artificial,gimeno2023artificial,intratorSelfsupervisedPolypReidentification2023,biffi_novel_2022}. These innovations promise to enhance quality control in clinical practice, improve cost-effectiveness, and streamline the assessment of endoscopist skills at individual, group, and large-scale levels~\cite{kaminski2017performance, tavanapong2022artificial, gimeno2023artificial}.

From a computer vision perspective, accurately listing polyps requires analyzing full-procedure colonoscopy videos, which can last from 20 minutes to over an hour. The challenge lies in detecting and re-identifying polyps (ReID) for accurate counting by correctly associating multiple tracklets—short, temporally consistent sequences of object detections—of the same polyp, without confusing them with others. Once the final set of tracklets is determined, the polyp count can be derived, and machine learning models can be applied to these tracklets for AI-based histopathological predictions~\cite{intratorSelfsupervisedPolypReidentification2023}. However, ReID is challenging due to polyps disappearing and reappearing, coexisting with others in the same frame, or changing appearance due to motion blur, lighting, camera distance, debris, and instruments~\cite{aizenman2024assessing,biffi2024real}. This field remains underexplored, with only the recent work by Intrator et al.~\cite{intratorSelfsupervisedPolypReidentification2023} framing polyp ReID as a two-stage approach: (1) polyp detection with tracking (e.g., ByteTrack), and (2) appearance-based re-identification, where SimCLR is employed to learn bounding box or tracklet representations. Tracklet similarity is then measured, and associations are established based on a predefined threshold.

We note that the task of re-identifying tracklets is similar to unsupervised person re-identification, where models must learn feature representations that are both discriminative, for distinguishing similar-looking individuals, and robust to variations in appearance caused by lighting, pose, or occlusion.
In this context, clustering algorithms, such as HDBSCAN~\cite{campello2013density}, are often used to group tracklets of the same individual based on visual similarity or assign pseudo-labels \cite{jahan2023unsupervised, zou2023discrepant}.
In this work, we hypothesize that unsupervised clustering algorithms using tracklet representations can significantly outperform the threshold-based approach proposed in \cite{intratorSelfsupervisedPolypReidentification2023}. Furthermore, the code and dataset from \cite{intratorSelfsupervisedPolypReidentification2023} remain private, limiting replication efforts. The recent release of the REAL-Colon dataset \cite{biffi2024real}, consisting of 60 annotated whole-procedure videos from multiple centers, offers an opportunity for benchmarking polyp counting approaches in an open-access setting.

\textbf{Contributions:}
This work makes four main contributions.
First, we re-implement and benchmark the method by Intrator et al.~\cite{intratorSelfsupervisedPolypReidentification2023} on the REAL-Colon dataset, detailing key adaptations to ensure the model functions correctly.
Second, we define a balanced data split for training and testing on the \realcolon{} dataset, enabling consistent evaluation across similar methods and providing robust metrics for the polyp counting task.
Third, we benchmark multiple clustering algorithms and propose an Affinity Propagation-based approach that substantially improves performance, establishing a new state-of-the-art in this field.
Lastly, we release data splits and code to reproduce all the results reported in this work.

\section{Method}
\label{sec:method}

\begin{figure*}[ht]
    \centering
    \begin{subfigure}[b]{0.53\textwidth}
        \centering
        \includegraphics[width=\textwidth]{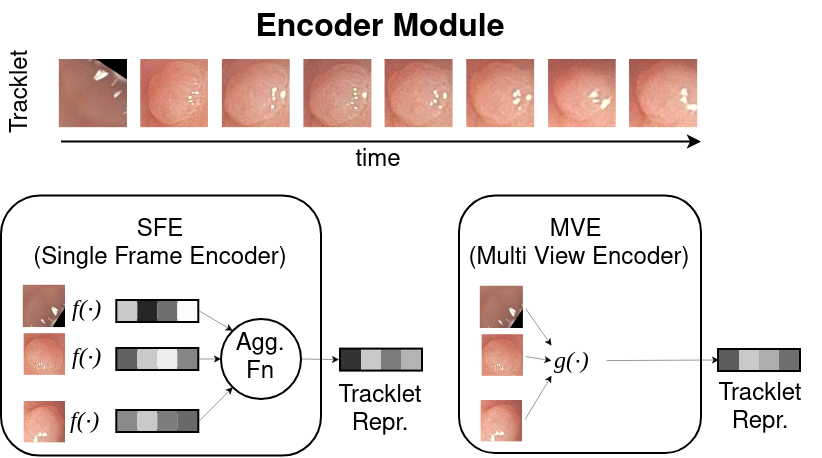}
        \caption{ }
        \label{fig:method-encoder}
    \end{subfigure}
    \hfill
    \begin{subfigure}[b]{0.45\textwidth}
        \centering
        \includegraphics[width=\textwidth]{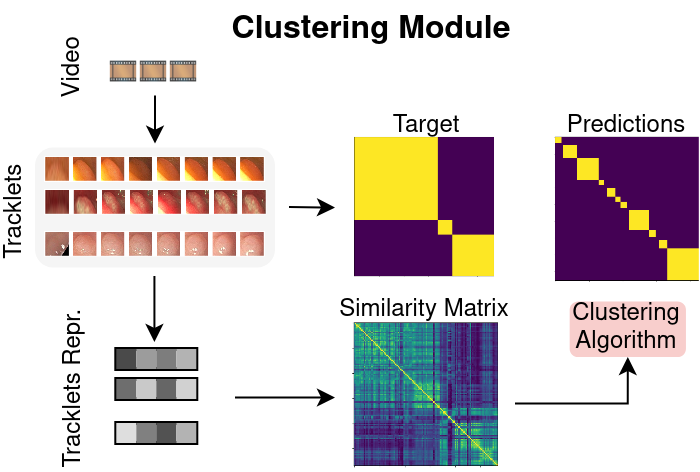}
        \caption{ }
        \label{fig:method-clustering}
    \end{subfigure}
    \caption{Proposed approach. \textbf{Fig.~\ref{fig:method-encoder}}. The encoder takes a polyp tracklet and computes its representation. \textbf{Fig.~\ref{fig:method-clustering}}. Pairwise similarities are computed from video tracklets and a clustering algorithm groups together trackelets to form polyp entities.}
    \label{fig:method}
\end{figure*}

In this work, we propose an approach for counting polyps in full-procedure videos, which consists of two main steps (Fig.~\ref{fig:method}).
First, following~\cite{intratorSelfsupervisedPolypReidentification2023}, a visual encoder obtains a tracklet representation from bounding boxes in single frames or in frame sequences (Fig.~\ref{fig:method-encoder}). Second, a clustering module re-associates these tracklets into polyp entities, leveraging visual cues extracted by the visual encoder (Fig.~\ref{fig:method-clustering}).

\subsection{Encoding polyp instances}
\label{sec:visual-encoder}

Intrator et al.~\cite{intratorSelfsupervisedPolypReidentification2023} showed that it is possible to learn features for polyp re-identification leveraging the SimCLR framework~\cite{chenSimpleFrameworkContrastive2020}.
SimCLR is a contrastive learning technique for visual representations.
It relies on image augmentations to generate a pair of positive samples in a self-supervised fashion.
However, since traditional image augmentations do not capture the diversity of polyp appearances in different contexts, the temporal nature of videos is leveraged to construct positive samples~\cite{intratorSelfsupervisedPolypReidentification2023}.
Authors propose two types of visual encoders, depicted in Fig.~\ref{fig:method-encoder}. The single-frame encoder (SFE) constructs the tracklet embedding by encoding and fusing together single frame representations. The multi-view encoder (MVE) is trained on sequences and learns tracklet representations end-to-end.

Inspired by findings in video representation learning~\cite{DBLP:conf/cvpr/QianMG0WBC21}, we modify the sampling such that it picks temporally close pair of images more frequently to learn temporally dependent features.
The SFE requires pairs of frames.
Let $l$ be the number of polyp instances for a given polyp entity.
We build a pair $(i, j)$ by uniformly picking $i$ in the available collection of frames $i = \mathcal{U}(1, l)$, while $j = \mathcal{G}(i, \sigma)$ is chosen from a Gaussian distribution centered on $i$.
The standard deviation $\sigma$ controls the frequency with which we sample frames closer to $i$.

The MVE works with pair of tracklets, i.e. sequences of frames instead of single frame.
In~\cite{intratorSelfsupervisedPolypReidentification2023}, a tracklet is split into three segments where the central one is discarded and the remaining fragments are collected to compose the positive pair.
However, this approach assumes that tracklets are short and confidently represent the entire polyp. In contrast, we utilize annotations that often result in longer tracklets, capturing the polyp at different stages of visibility (e.g. early or late detections).
To address this, we sample the initial frame for the first and second fragments of each pair from their corresponding halves within the tracklet. The remaining frames are picked with a stride in $\{1, 2, 3, 4\}$ for each fragment, enhancing diversity in frame selection and ensuring a broader representation of the polyp.

To learn robust representation of polyps, we introduce another sampling strategy where the two fragments are gathered from two different tracklets $t_1$, $t_2$ belonging to the same polyp entity.
Fragments from different tracklets may differ significantly in appearance, leading to noisy training. To address this, inspired by curriculum learning~\cite{DBLP:conf/icml/BengioLCW09}, we gradually incorporate such pairs into training by increasing their sampling probability in later training stages.

\subsection{Re-association via clustering}
\label{sec:clustering}

The goal of the clustering module (Fig.~\ref{fig:method-clustering}) is to group together tracklets belonging to the same polyp based on visual appearance. Noticing the similarity with person re-identification task, we turn to unsupervised clustering to solve re-association and rephrase it as a clustering problem.

Given a tracklet $t = [1, 2, \ldots, n] \in T$ as a list of frame indices with $[x_1, x_2, \ldots, x_n]$ the corresponding frames, we obtain its embedding $e_t$ through the visual encoder.
Given the single-frame encoder $f$, frame representations are fused together with an aggregation function like the mean: $e_t = \sum_i^n f(x_i) / n$, while the multi-view encoder directly return the tracklet's representation $e_t = g(t)$.
Then, given a set of tracklets $T$, we compute the embedding for every tracklet $\{e_1, e_2, \ldots, e_n\}$ and construct a pairwise distance matrix $D \in \mathbb{R}^{n \times n}$, $D_{i,j} = \text{dist}(e_{t_i}, e_{t_j})$ where $\text{dist}$ is a distance function i.e. the Euclidean distance. This matrix represent how distant (i.e. how dissimilar) a pair of tracklet is.
We then obtain a similarity matrix $S \in \mathbb{R}^{n \times n}$ by inverting and normalizing the pairwise distances: $S = 1 - \frac{D - \min(D)}{\max(D) - \min(D)}$, where each $S_{i,j} \in [0, 1]$ and similar tracklets present higher scores.

Given a tracklet $t_i$ and the similarity matrix $S$, current literature re-associates $t_i$ with every tracklet $t_j$ such that $S_{i,j} \geq \lambda$, where $\lambda \in [0,1]$ is a threshold~\cite{intratorSelfsupervisedPolypReidentification2023}.
We argue that this threshold-based clustering can be replaced by more effective clustering algorithms, without requiring further training data.
We tested three unsupervised clustering methods: hierarchical clustering~\cite{DBLP:journals/widm/MurtaghC12}, which iteratively merges or splits data points based on distance metrics to build a hierarchy of clusters; HDBSCAN~\cite{campello2013density}, a density-based clustering method that extends DBSCAN by extracting clusters of varying densities without predefining cluster count; and affinity propagation~\cite{frey2007clustering}, which selects representative samples (exemplars) for clusters by iteratively passing similarity messages between data points.

\subsection{Evaluation framework}
\label{sec:metrics}

In~\cite{intratorSelfsupervisedPolypReidentification2023}, authors propose to evaluate ReID effectiveness using the average polyp fragmentation rate (FR), defined as the average number of tracklets polyps are split into. The FR is a number $\geq 1$, where lower values mean lower fragmentation.
Formally, we define the FR as a function of $T$ a set of tracklets and $E$ a set of polyp entities. Given a set of re-associated tracklets $R$, with $|R| \leq |T|$, we can compute the fragmentation rate as $FR = |R| / |E|$. A perfect re-association would have $|R| = |E|$, thus $FR = 1$.

FR successfully evaluates effectiveness in re-association but does not account for wrong tracklet matching, i.e. association to the wrong polyp entity (false positive).
For this reason, following~\cite{intratorSelfsupervisedPolypReidentification2023} we measure the fragmentation rate at a specific rate of false positive tracklet re-associations $\rho$.
In our application, we set $\rho = 0.05$, i.e. 5\% false positive rate.
Thus, $FR = |\hat{R}| / |E|$, where $\hat{R}$ is the set of re-associated tracklets that minimizes $|FPR(R) - \rho|$.
The $FPR$ function returns the rate of false positives for the set of re-associated tracklets $R$ by matching them to targets.
While~\cite{intratorSelfsupervisedPolypReidentification2023} does not specify the approach, we select $\hat{R}$ by tuning the clustering algorithm's hyper-parameters on the validation set.

We compute the FR intra-video, for each video in the collection and then (macro-)average results.
We chose to average over videos instead of tracklets to keep equal weights between different videos and ensure results are not disproportionately influenced by videos with a higher number of tracklets.

\section{Experiments}
\label{sec:experiments}

\subsection{Dataset}
\label{sec:exp-setup}

We use the \realcolon{} dataset in our experiments. It includes 60 annotated videos of full-length colonoscopies among 4 cohorts~\cite{biffi2024real}.
In~\cite{aizenman2024assessing} authors propose a split on \realcolon{} to evaluate polyp detection clinical efficacy. Similarly, we propose a split of \realcolon{} that enables a comprehensive evaluation of polyp counting on full-length procedures.

This allocation between evaluation splits is carefully balanced to reflect key factors such as the total number of polyps and videos in each split, the presence of single-polyp videos, and the distribution of videos across cohorts. We split \realcolon{} into three sets. The training set is built from the first 8 videos per cohort, totaling 206110 annotated frames. The validation and test sets are carefully selected from the remaining 7 videos per cohort, totaling 145150 frames. Table~\ref{tab:split} reports the properties of above mentioned splits analytically.

\begin{table}[hb]
\centering
\begin{tabular}{l | C{1.0cm} C{1.7cm} C{1.7cm} C{1.6cm}}
Split & n. polyps & n. vids with $=$ 1 polyp & n. vids with $\geq$ 2 polyps & n. vids per cohort \\ \toprule
train & 86 & 13 & 19 & 6/7/8/6 \\
val & 22 & 5 & 5 & 2/2/3/3 \\
test & 24 & 4 & 5 & 1/2/3/3 \\
\end{tabular}
\caption{Properties of the proposed train, validation and test sets on \realcolon{}. Number is abbreviated with ``n.''. The number of videos per cohort follows the format cohort1/cohort2/cohort3/cohort4. }
\label{tab:split}
\end{table}

\subsection{Implementation details}
\label{sec:impl}

The visual encoders are implemented following~\cite{intratorSelfsupervisedPolypReidentification2023}.
During training, we set single-view's $\sigma = 30$ and batch size $= 64$.
We use standard image augmentations, i.e. horizontal and vertical flipping, affine augmentations (rotation 15 degrees, translation 10\%, scaling 10\%) and color jittering (brightness 0.2, contrast 0.2, saturation 0.2, hue 0.1).
Every frame is cropped to the polyp instance bounding box and resized to 232x232.
Batch size is set to 16 for the multi-view encoder.
To avoid introducing noise from an external tracker and to establish a common detection baseline for this and future approaches, we construct tracklets directly from the provided annotations instead of employing an off-the-shelf tracker like ByteTrack~\cite{zhangByteTrackMultiobjectTracking2022}.
A tracklet is defined as a sequence of consecutive frames from the same polyp entity, with an Intersection Over Union (IoU) of at least $0.1$ between consecutive frames.
Following~\cite{intratorSelfsupervisedPolypReidentification2023}, each tracklet is split into fragments of 8 frames, due to memory limitations.
The same augmentations are applied on fragments.
A linear scheduler controls the probability $\alpha$ of sampling fragments from different tracklets. Initially, $\alpha = 1$ and gradually decreases to $0.5$ over the first 75\% of training.
A batch is created by selecting fragments from different polyps, unlike~\cite{intratorSelfsupervisedPolypReidentification2023}, where pairs came from different procedures, as we have a much lower number of procedures available.
The best visual encoder was chosen evaluating top-1 accuracy~\cite{chenSimpleFrameworkContrastive2020} on the validation set using $\sigma = 30$ and $\alpha = 0.5$.
During evaluation, we obtain full tracklet representation by encoding one frame every $s = 4$.
The target false positive rate $\rho$ is $0.05$, meaning that we allow for 5\% of false positive re-associations.

\subsection{Results}

We compare our approach on \realcolon{} to the state-of-the-art~\cite{intratorSelfsupervisedPolypReidentification2023} and present results in Table~\ref{tab:results}.
Our method obtains best results with the multi-view encoder and the affinity propagation algorithm for tracklet re-association.
The clustering algorithm and its hyper-parameters are selected on the validation set with a target false positive rate (FPR) of 5\%.
Our method generalizes over the test set and consistently maintains an FPR of 5\% or lower, unlike the threshold-based clustering proposed in~\cite{intratorSelfsupervisedPolypReidentification2023}.
Furthermore, results on \realcolon{} test set highlight that our method is able to fragment a polyp 3.9 times less compared to previous state-of-the-art.
On average, our re-association approach splits a polyp entity into 6.30 fragments, while threshold-based clustering into 24.60 and 27.90 fragments.
These results demonstrate our method's robust performance on full-procedure colonoscopy videos. We note that~\cite{intratorSelfsupervisedPolypReidentification2023} reports lower fragmentation rates on their proprietary test set. The reason is twofold. First, \realcolon{} starts from more fragmented tracklets, making the task harder: we report an initial fragmentation rate of 56.33 with respect to their 3.30. Moreover, we train the model on a low data regime—32 videos—while they have access to nearly 15k procedures.

\begin{table}
\centering
\footnotesize
\begin{tabular}{l|ccc|ccc}
\toprule
\multirow{2}{*}{Method} & \multicolumn{3}{c|}{Validation set} & \multicolumn{3}{c}{Test set} \\
 & FR & FR$_{std}$ & FPR & FR & FR$_{std}$ & FPR \\ \midrule 
No ReID & 35.63 & 18.60 & - & 56.33 & 45.53 & - \\ 
SFE~\cite{intratorSelfsupervisedPolypReidentification2023} & 18.25 & 11.10 & 0.0504 & 24.60 & 21.29 & 0.1276 \\ 
MVE~\cite{intratorSelfsupervisedPolypReidentification2023} & 17.30 & 11.14 & 0.0464 & 27.90 & 21.43 & 0.1030 \\
\textbf{Ours} & \textbf{4.60} & \textbf{2.04} & \textbf{0.0681} & \textbf{6.30} & \textbf{4.47} & \textbf{0.0431} \\
\bottomrule
\end{tabular}
\caption{Fragmentation rate results on \realcolon{}. The FR and FR$_{std}$ column report the average fragmentation rate and its standard deviation, lower is better. The FPR column reports the rate of false positive matches. Test set's results are obtained using best hyper-parameters selected on the validation set targeting 0.05 FPR.}
\label{tab:results}
\end{table}

\subsection{Ablation}

We evaluate the impact of different clustering algorithms on polyp counting, comparing results from both single-view and multi-view encoders. Table~\ref{tab:ablation} presents these results.
The threshold-based algorithm from~\cite{intratorSelfsupervisedPolypReidentification2023} results in higher false positive rates on the test set.
While generally providing better results, the multi-view encoder offers only limited improvement over the single-view encoder on the \realcolon{} benchmark.
This is likely due to the Transformer used for sequence encoding, which is known to be data hungry~\cite{DBLP:conf/iclr/DosovitskiyB0WZ21}. However, in~\cite{intratorSelfsupervisedPolypReidentification2023}, authors demonstrated that substantially larger datasets—they work with more than 15k videos—can further enhance MVE performance, yet such expansive data resources are often unavailable.
On the contrary, our proposed clustering algorithms that do not require additional training data.
On this open-access benchmark, our approach improves the performance while maintaining adaptability to smaller datasets.

\begingroup
\setlength{\tabcolsep}{2.0pt}
\begin{table}
\centering
\footnotesize
\begin{tabular}{l|ccc|ccc}
\toprule
\multirow{2}{*}{Method} & \multicolumn{3}{c|}{Validation set} & \multicolumn{3}{c}{Test set} \\
 & FR & FR$_{std}$ & FPR & FR & FR$_{std}$ & FPR \\
\midrule
No ReID & 35.63 & 18.60 & - & 56.33 & 45.53 & - \\ \hline
\multicolumn{7}{l}{\textbf{\textit{Single-frame encoder}}} \\
Threshold~\cite{intratorSelfsupervisedPolypReidentification2023} & 18.25 & 11.10 & 0.0504 & 24.60 & 21.29 & 0.1276 \\
\textit{Ours} w/ Agglomerative~\cite{DBLP:journals/widm/MurtaghC12} & 7.53 & 4.33 & 0.0499 & 14.88 & 13.98 & 0.0249 \\
\textit{Ours} w/ HDBSCAN~\cite{campello2013density} & 6.78 & 5.03 & 0.0428 & 15.24 & 13.26 & 0.0476 \\
\textit{Ours} w/ Affinity Prop.~\cite{frey2007clustering} & \textbf{5.03} & \textbf{2.52} & \textbf{0.0491} & \textbf{6.78} & \textbf{5.03} & \textbf{0.0428} \\
\hline
\multicolumn{7}{l}{\textbf{\textit{Multi-view encoder}}} \\
Threshold~\cite{intratorSelfsupervisedPolypReidentification2023} & 17.30 & 11.14 & 0.0464 & 27.90 & 21.43 & 0.1030 \\
\textit{Ours} w/ Agglomerative~\cite{DBLP:journals/widm/MurtaghC12} & 10.63 & 7.76 & 0.0517 & 21.83 & 20.52 & 0.0236 \\
\textit{Ours} w/ HDBSCAN~\cite{campello2013density} & 8.92 & 4.62 & 0.0515 & 13.57 & 10.11 & 0.0356 \\
\textit{Ours} w/ Affinity Prop.~\cite{frey2007clustering} & \textbf{4.60} & \textbf{2.04} & \textbf{0.0681} & \textbf{6.30} & \textbf{4.47} & \textbf{0.0431} \\
\bottomrule
\end{tabular}
\caption{Results for different clustering algorithms on \realcolon{}. Notation follows Table~\ref{tab:results}.}
\label{tab:ablation}
\end{table}
\endgroup

\section{Conclusion}
\label{sec:conclusion}
In conclusion, this is the first study to tackle the problem of polyp counting in full-procedure videos in a fully open-access setting. We demonstrate how the REAL-Colon dataset can be effectively utilized for this task and re-implement and adapt existing polyp ReID methods for this goal. Furthermore, we propose an unsupervised clustering approach that significantly reduces fragmentation by 3.9 times, setting a new baseline for this problem. We believe that this work, along with the experimental framework, will foster future research on automated colonoscopy reporting. To support this, we release data splits and the codebase for result replication.


\section{Compliance with ethical standards}

This research study was conducted retrospectively using human subject data made available in open access by~\cite{biffi2024real}. Ethical approval was not required as confirmed by the license attached with the open-access data.

\section{Competing interests}

C.B. and A.C. are affiliated with Cosmo Intelligent Medical Devices, the developer of the GI Genius medical device.

\bibliographystyle{IEEEbib}
\bibliography{refs}

\begin{thebibliography}{10}

\bibitem{berzin_position_2020}
T.~M. Berzin et~al.,
\newblock ``Position statement on priorities for artificial intelligence in gi endoscopy: a report by the asge task force,''
\newblock {\em Gastrointest. Endosc.}, vol. 92, pp. 951--959, 2020.

\bibitem{spadaccini_computer-aided_2021}
M.~Spadaccini et~al.,
\newblock ``Computer-aided detection versus advanced imaging for detection of colorectal neoplasia: a systematic review and network meta-analysis,''
\newblock {\em Lancet Gastroenterol. Hepatol.}, vol. 6, pp. 793--802, 2021.

\bibitem{aizenman2024assessing}
Gabriel Aizenman, Pietro Salvagnini, Andrea Cherubini, and Carlo Biffi,
\newblock ``Assessing clinical efficacy of polyp detection models using open-access datasets,''
\newblock {\em Frontiers in oncology}, vol. 14, pp. 1422942, 2024.

\bibitem{biffi_novel_2022}
Carlo Biffi, Pietro Salvagnini, Nhan Ngo~Dinh, Cesare Hassan, Prateek Sharma, {GI Genius CADx Study Group}, and Andrea Cherubini,
\newblock ``A novel {AI} device for real-time optical characterization of colorectal polyps,''
\newblock {\em NPJ digital medicine}, vol. 5, no. 1, pp. 84, June 2022.

\bibitem{tavanapong2022artificial}
Wallapak Tavanapong, JungHwan Oh, Michael~A Riegler, Mohammed Khaleel, Bhuvan Mittal, and Piet~C De~Groen,
\newblock ``Artificial intelligence for colonoscopy: Past, present, and future,''
\newblock {\em IEEE journal of biomedical and health informatics}, vol. 26, no. 8, pp. 3950--3965, 2022.

\bibitem{gimeno2023artificial}
Antonio~Z Gimeno-Garc{\'\i}a, Anjara Hern{\'a}ndez-P{\'e}rez, David Nicol{\'a}s-P{\'e}rez, and Manuel Hern{\'a}ndez-Guerra,
\newblock ``Artificial intelligence applied to colonoscopy: Is it time to take a step forward?,''
\newblock {\em Cancers}, vol. 15, no. 8, pp. 2193, 2023.

\bibitem{intratorSelfsupervisedPolypReidentification2023}
Yotam Intrator, Natalie Aizenberg, Amir Livne, Ehud Rivlin, and Roman Goldenberg,
\newblock ``Self-supervised {{Polyp Re-identification}} in~{{Colonoscopy}},''
\newblock in {\em Medical {{Image Computing}} and {{Computer Assisted Intervention}}}, 2023, pp. 590--600.

\bibitem{kaminski2017performance}
Michal~F Kaminski et~al.,
\newblock ``Performance measures for lower gastrointestinal endoscopy: a european society of gastrointestinal endoscopy (esge) quality improvement initiative,''
\newblock {\em Endoscopy}, vol. 49, no. 04, pp. 378--397, 2017.

\bibitem{biffi2024real}
Carlo Biffi, Giulio Antonelli, Sebastian Bernhofer, Cesare Hassan, Daizen Hirata, Mineo Iwatate, Andreas Maieron, Pietro Salvagnini, and Andrea Cherubini,
\newblock ``Real-colon: A dataset for developing real-world ai applications in colonoscopy,''
\newblock {\em Scientific Data}, vol. 11, no. 1, pp. 539, 2024.

\bibitem{campello2013density}
Ricardo~JGB Campello, Davoud Moulavi, and J{\"o}rg Sander,
\newblock ``Density-based clustering based on hierarchical density estimates,''
\newblock in {\em Pacific-Asia conference on knowledge discovery and data mining}, 2013, pp. 160--172.

\bibitem{jahan2023unsupervised}
Meskat Jahan, Manajir Hassan, Sahadat Hossin, Md~Iftekhar Hossain, and Mahmudul Hasan,
\newblock ``Unsupervised person re-identification: A review of recent works,''
\newblock {\em Neurocomputing}, p. 127193, 2023.

\bibitem{zou2023discrepant}
Chang Zou, Zeqi Chen, Zhichao Cui, Yuehu Liu, and Chi Zhang,
\newblock ``Discrepant and multi-instance proxies for unsupervised person re-identification,''
\newblock in {\em International Conference on Computer Vision}, 2023, pp. 11058--11068.

\bibitem{chenSimpleFrameworkContrastive2020}
Ting Chen, Simon Kornblith, Mohammad Norouzi, and Geoffrey Hinton,
\newblock ``A {{Simple Framework}} for {{Contrastive Learning}} of {{Visual Representations}},''
\newblock in {\em {{International Conference}} on {{Machine Learning}}}, 2020, pp. 1597--1607.

\bibitem{DBLP:conf/cvpr/QianMG0WBC21}
Rui Qian et~al.,
\newblock ``Spatiotemporal contrastive video representation learning,''
\newblock in {\em Computer Vision and Pattern Recognition}, 2021, pp. 6964--6974.

\bibitem{DBLP:conf/icml/BengioLCW09}
Yoshua Bengio, Jérôme Louradour, Ronan Collobert, and Jason Weston,
\newblock ``Curriculum learning,''
\newblock in {\em International Conference on Machine Learning}, 2009, vol. 382, pp. 41--48.

\bibitem{DBLP:journals/widm/MurtaghC12}
Fionn Murtagh and Pedro Contreras,
\newblock ``Algorithms for hierarchical clustering: An overview,''
\newblock {\em WIREs Data Mining Knowl. Discov.}, vol. 2, no. 1, pp. 86--97, 2012.

\bibitem{frey2007clustering}
Brendan~J Frey and Delbert Dueck,
\newblock ``Clustering by passing messages between data points,''
\newblock {\em science}, vol. 315, no. 5814, pp. 972--976, 2007.

\bibitem{zhangByteTrackMultiobjectTracking2022}
Yifu Zhang, Peize Sun, Yi~Jiang, Dongdong Yu, Fucheng Weng, Zehuan Yuan, Ping Luo, Wenyu Liu, and Xinggang Wang,
\newblock ``{{ByteTrack}}: {{Multi-object Tracking}} by {{Associating Every Detection Box}},''
\newblock in {\em European Conference on Computer Vision}. 2022.

\bibitem{DBLP:conf/iclr/DosovitskiyB0WZ21}
Alexey Dosovitskiy et~al.,
\newblock ``An image is worth 16x16 words: Transformers for image recognition at scale,''
\newblock in {\em International Conference on Learning Representations}, 2021.

\end{thebibliography}

\end{document}